\documentclass{article}
\usepackage{amssymb}
\usepackage{graphicx}
\usepackage{graphicx}
\usepackage{hyperref}
\usepackage{xcolor}
\usepackage{amsmath}
\usepackage{enumitem}
\usepackage{graphics}
\usepackage{graphicx}
\usepackage{float}
\usepackage{url}
\usepackage{subcaption}
\usepackage{mwe}
\usepackage{authblk}

\title{Modelling SARS-CoV-2 coevolution with genetic algorithms}
\author[1,2]{Aymeric Vi\'{e}}
\affil[1]{Mathematical Institute, University of Oxford}
\affil[2]{Institute of New Economic Thinking, University of Oxford}
\date{\today}
\begin{document}

\maketitle

\begin{abstract}
At the end of 2020, policy responses to the SARS-CoV-2 outbreak have been shaken by the emergence of virus variants, impacting public health and policy measures worldwide.
The emergence of these strains suspected to be more contagious, more severe, or even resistant to antibodies and vaccines, seem to have taken by surprise health services and policymakers, struggling to adapt to the new variants constraints. 
Anticipating the emergence of these mutations to plan ahead adequate policies, and understanding how human behaviors may affect the evolution of viruses by coevolution, are key challenges. 
In this article, we propose coevolution with genetic algorithms (GAs) as a credible approach to model this relationship, highlighting its implications, potential and challenges. 
Because of their qualities of exploration of large spaces of possible solutions, capacity to generate novelty, and natural genetic focus, GAs are relevant for this issue. We present a dual GA model in which both viruses aiming for survival and policy measures aiming at minimising infection rates in the population, competitively evolve.
This artificial coevolution system may offer us a laboratory to "debug" our current policy measures, identify the weaknesses of our current strategies, and anticipate the evolution of the virus to plan ahead relevant policies. It also constitutes a decisive opportunity to develop new genetic algorithms capable of simulating much more complex objects. We highlight some structural innovations for GAs for that virus evolution context that may carry promising developments in evolutionary computation, artificial life and AI. 
\end{abstract}

\section{Introduction}

\begin{figure}[H]
    \centering
    \includegraphics[width = .8\textwidth]{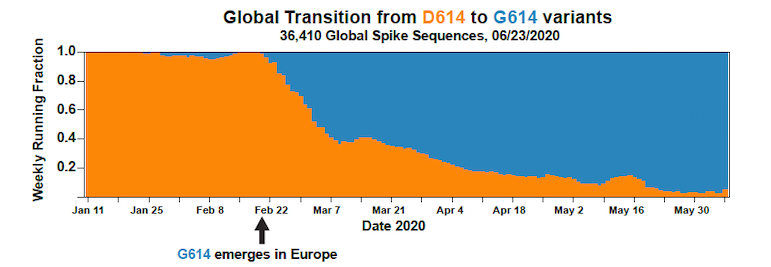}
    \caption{Shift over time from orange (the original D type of the virus) to blue (the now-widespread G form, D614G); (Los Alamos National Laboratory, 2020)}
    \label{timeline}
\end{figure}

As early as June 2020, the initial SARS-CoV-2 strain identified in China was replaced as the dominant variant by the D614G mutation (Figure \ref{timeline}). Appeared in January 2020, this strain differed because of a substitution in the gene encoding the spike protein. The D614G substitution has been found to have increased infectivity and transmission (WHO, 2020a; Korber et al., 2020).\par

On November 5 2020, a new strain of SARS-CoV-2 was reported in Denmark (WHO, 2020b), linked with the mink industry. The "unique" mutations identified in one cluster, "Cluster 5", seemingly as contagious or severe as others, has been found to moderately decrease the sensitivity of the disease to neutralising antibodies. Culling of farmed minks, increase of genome sequencing activities and numerous closing of borders to Denmark residents, followed.\par
On 14 December 2020, the United Kingdom reported a new variant VOC 202012/01, with a remarkable number of 23 mutations, with unclear origin (Kupferschmidt, 2020). Early analyses have found that the variant has increased transmissibility, though no change in disease severity was identified (WHO, 2020a). One of these 23 mutations, the deletion at position 69/70del, was found to affect the performance of some PCR tests, currently at the center of national testing strategies. Quickly becoming dominant, this variant was held responsible for a significant increase in mortality, ICU occupation and infections across the country (Iacobucci, 2021; Wallace and Ackland, 2021). \par
On 18 December, the variant 501Y.V2 was detected in South Africa, after rapidly displacing other virus lineages in the region. Preliminary studies showed that this variant was associated with a higher viral load, which may cause increased transmissibility (WHO, 2020a). Recent findings have shown that this variant significantly reduced the efficacy of vaccines (Mahase, 2021). \\

RNA viruses have high mutation rates (Duffy, 2018). Although many mutations are not beneficial for the organisms, and some are inconsequential, some small fraction of them are beneficial. We refer the reader to (Duffy, 2018) and Domingo et al. (1996) for a discussion on RNA viruses mutation rates. The consequences of these high mutation rates notably are higher evolvability, i.e. higher capacity to adapt to changing environments. This allows them to emerge in new hosts, escape vaccine-induced immunity, or circumvent disease resistance. However, RNA viruses seem to be just below the threshold for critical error: if the majority of mutations are deleterious, higher mutation rates may cause ecological collapse in the virus population. As a RNA virus (Lima, 2020), SARS-CoV-2 shares these characteristics, and mutates very frequently (Phan, 2020; Benvenuto et al. (2020), Maty{\'{a}}sek, and Kovar{\'{i}}k, 2020). Especially relevant for this class of virus, the priorities of many researchers including the WHO Virus Evolution Working Group, have been to strengthen ways to identify relevant mutations, study their characteristics and impacts, as well as outlining mitigation strategies to respond to these mutations (WHO, 2020a). 

Anticipating the emergence of these mutations to plan ahead adequate policies, and understanding how human behaviors may affect the evolution of viruses by coevolution, are key challenges. Human adaptation of policies and behaviors can impact the reproduction of SARS-CoV-2, and target specific characteristics such as airborne transmission. The impact of human policies and behaviors on outbreak trajectory, the evaluation of non pharmaceutical measures, have been the object of numerous analyses. However, most of these analyses do not include the possibility for viruses to mutate, with novel effects and increased transmission rates. The space of possible virus strains is huge and to some extent quasi open-ended, challenging modelling attempts of this arms' race.

In this article, we propose coevolution with genetic algorithms (GAs) as a credible approach to model this relationship, highlighting its implications, potential and challenges. We provide a proof of concept-implementation of this coevolution dual-GA.  Because of their qualities of exploration of large spaces of possible solutions, capacity to generate novelty, and natural genetic focus, GAs are relevant for this issue. We present a dual GA model in which both viruses aiming for survival and policy measures aiming at minimising infection rates in the population, competitively evolve. Under coevolution, virus adaptation towards more infectious variants appear considerably faster than when the virus evolves against a static policy. More contagious strains become dominant in the virus population under coevolution. The coevolution regime can generate multiple outbreaks waves as the more infectious variants becoming more dominant in the virus population. Seeing  more  infectious  virus  variants  becoming  dominants  may signify  that  our  policy  measures  are  effective.  

This artificial coevolution system may offer us a laboratory to "debug" our current policy measures, identify the weaknesses of our current strategies, and anticipate the evolution of the virus to plan ahead relevant policies. It highlights how human behaviors can shape the evolution of the virus, and how reciprocally the evolution of the virus shapes the adaptation of public policy measures. To overcome the simplifications of the implementation in this article, several key innovations for evolutionary algorithms may be required, in particular bringing more advanced biological and genetic concepts in current evolutionary algorithms. 

We first present in Section \ref{coevolution_section} the concept of coevolution, both generally in complex systems, and specifically in our study of the evolution viruses and policies. We propose genetic algorithms as a modelling tool for this context. Genetic algorithms are briefly introduced in Section \ref{GA_section}. We present our perspective of using genetic ,algorithms to generate an artificial coevolution of SARS-CoV-2, and present its main concepts and design in Section \ref{description}. Then, we propose an example of implementation of a dual genetic algorithm to model this coevolution process in Section \ref{example}, describing the model, the operators, the parameters, and some key results. We develop further the implications and perspectives of this work in Section \ref{perspectives}. Section \ref{data} presents data and code availability, and Section \ref{conclusion} concludes.

\section{Coevolution of virus traits and policy actions}
\label{coevolution_section}

\subsection{Coevolution in complex systems}

Co-evolution opens a promising and new way to model such ecosystems. Investors in the stock market evolve financial strategies to obtain higher profit, and this evolution can be captured by a GA model. But they are evolving in an environment, that notably includes financial regulations set by policy makers. Not only these regulations are evolving as policy makers strive to identify the best policy to stabilise the market and avoid large crashes: the evolution of regulation and financial strategies is a co-evolution of two species. Policy makers attempt to discourage new loopholes exploited by investors that set a threat on the real economy; investors adapt to the new regulations seeking for other ways to extract profit, finding new niches that trigger new adaptations of regulations. By capturing this interplay, a GA approach could act as a \textit{debugging tool} for financial regulations, a \textit{stress-test program} that invents novel ways to challenge our organisations.  \par
Most sports competitions see such interplay between rules and strategies. The 2008 Olympic Games saw controversy over new swimming suits with novel materials that allowed unprecedented speed and records, leading to their ban causing a change in the innovation strategies of manufacturers. This new direction may spark some day a similar story, calling for new regulation, sparking a different evolution trajectory. Formula 1 constructors actively seek grey-area zones in the regulation hoping for marginal performance gains. One team creatively bypassed the action of a regulatory sensor to increase its engine power, pushing the regulations to add a second sensor and regulate the use of engine modes, impacting all teams' performance. Another racing team exploited unclear rules on purchases and copying of other cars' parts to, leading to a change in the regulations that impacts the evolution of other teams development programs, and that may as well create further unclear rules to be abused in the future. Another instance of coevolution in complex systems, of public high interest, is the co-evolution of viruses and population behaviors or policy measures. 

\subsection{The coevolution of SARS-CoV-2 and policy measures}

\begin{figure}[H]
    \centering
    \includegraphics[width = .8\textwidth]{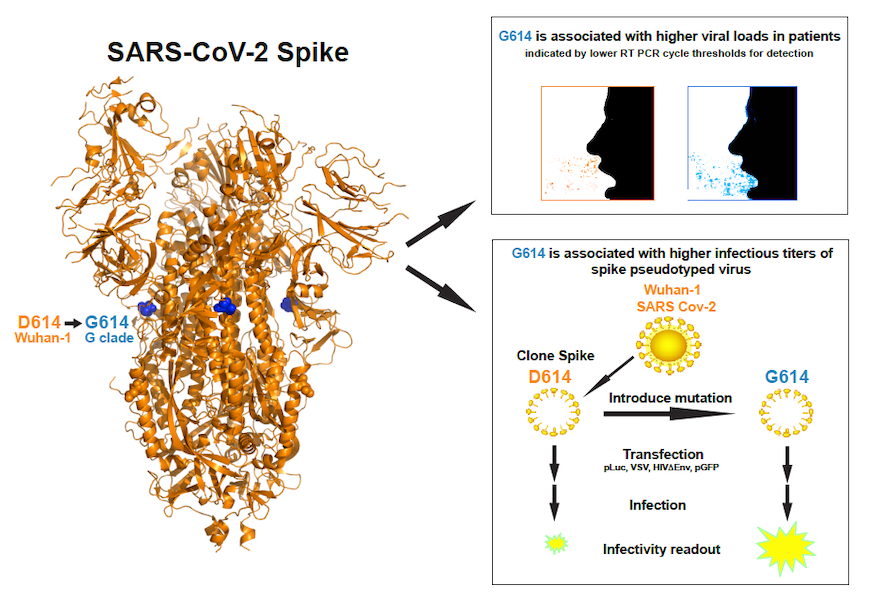}
    \caption{Illustration of the mutation leading to the variant D614G; (Los Alamos National Laboratory, 2020)}
    \label{spike}
\end{figure}

The emergence of viruses' mutations is a complex topic, both in the mechanisms involved at the virus genome level, but also on what causes some particular mutations to appear, or to be rewarded. That is, the fitness (dis)advantage of the new trait encoded by a mutation, in its environment. We can see the struggle between SARS-CoV-2 mutations illustrated in Figure \ref{spike}, and human behaviors and policy measures, as an arms race, a coevolution. Humans adopt new restrictions, wear face coverings, adopt social distancing measures, develop testing methods, to reduce the fatalities and infections due to the virus. Facing this pressure, the virus' mutations unconsciously strive to change its genome in order to improve its chances of survival. As some mutations allow the virus to get new, beneficial traits, possibly higher transmissibility (Priya and Shanker, 2021), resistance to antibodies (Callaway, 2020) or causing anomalies in PCR tests (WHO, 2020a), human behaviors may adapt, continuing the arms race. This evolutionary change in traits of individuals in one population, in response to a change of trait in a second population, followed by a reciprocal response, is a phenomenon known as coevolution (Janzen, 1980). Viruses are walking on the fitness landscape (Wright, 1931), a physical representation of the relationship between traits and fitness, and humans change by their behavior this fitness landscape. If by example all humans were hypothetically wearing perfectly hermetic face coverings, airborne transmission methods would fail, causing the virus either to go extinct, or to find other means of transmission. \\
    
The continuous interplay between individual genomes or characteristics, and their environment, is an endless source of novelty and niches for adaptation. Individuals are influenced by their environment, and the environment itself is influenced by individual. This dynamic is difficult to model, especially in our context of virus and policies coevolution. The space of possible actions or policy measures is at least very large. Humans can adopt a large diversity of measures, with many levels of stringency or public support. Likewise, the large size of the space of possible genomes for viruses, and the diversity of phenotypes, i.e. observable characteristics, that they can exhibit, challenge our modelling attempts. Coevolution can give birth to novel traits that did not exist before, in a quasi open-ended process. Random or enumerative search methods struggle to evaluate such a large number of possible combinations. We propose here an alternative framework to simulate this coevolution phenomenon in spite of the complexity of the task. Modelling coevolutionary dynamics has seen a large variety of approaches: stochastic processes mathematical modelling (Dieckmann and Law, 1996, Hui et al., 2018), network science (Guimaraes et al., 2017), dynamical systems (Caldarelli et al., 1998), and more biological or genetic methods (Gilman et al., 2012). Evolutionary algorithms (EAs, used for coevolution with Rosin and Belew, 1997), in particular Genetic algorithms (GAs), offer one promising approach at this end. Let us first introduce them briefly, before outlining the properties that makes them relevant for this task.

\section{Evolutionary and Genetic Algorithms}
\label{GA_section}
    
A genetic algorithm (GA) is a member of the family of \textit{evolutionary algorithms} (EAs), that are computational search methods inspired from natural selection (Holland, 1992). They simulate Darwinian evolution on individual entities, gathered in a \textit{population}. Genetic algorithms represent these entities with a \textit{genome}, i.e. a collection of genes, often represented as a bit string, that determines the entity \textit{phenotype}, i.e. observable characteristics. The entities undergo selection based of fitness, reproduction of fittest entities, mutations of the genome, that affect their traits (Mirjalili, 2019). Iterating this simplified evolution process, the characteristics of the entities may change, improving the fitness of the population. \\
    
As a population-based search method, GAs are efficient in the exploration of \textit{search spaces}, i.e. space of possible solutions, that can be very large (Axelrod, 1987), or rugged (Wiransky, 2020). That is, that admit several extrema, or very irregular structure. They quickly identify regions of the search space that are associated with higher fitness, showing satisfying optimisation capacities (Bhandari et al., 1996). They can also be used to model evolutionary systems, from economies and financial strategies to biological ecologies. Vie (2020a) reviews in more detail its qualities and perspectives as a search method and a modelling tool.
    
\section{An artificial coevolution of SARS-CoV-2}
\label{description}
    
\begin{figure}[H]
\centering
\includegraphics[width=\textwidth]{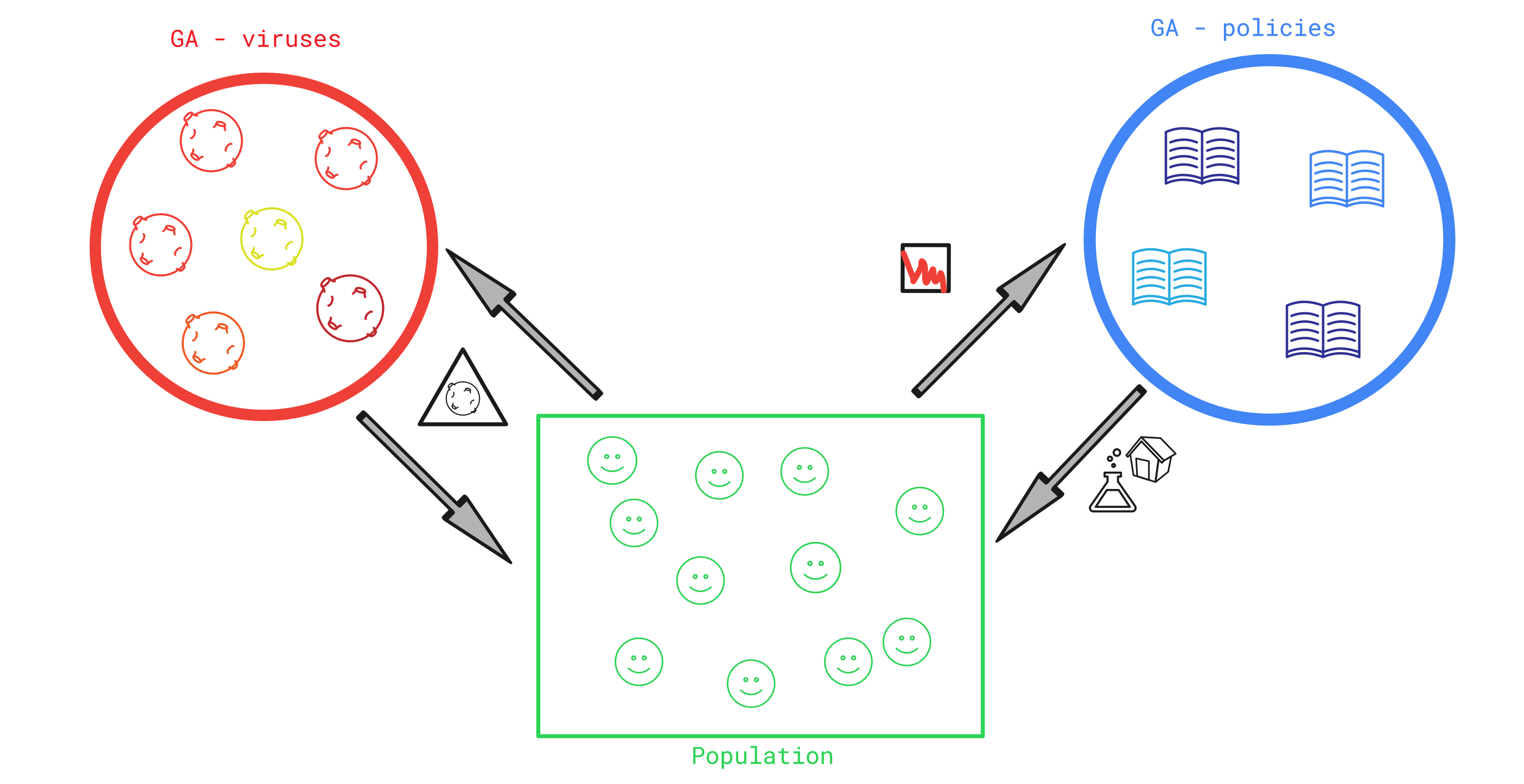}
\caption{The coevolution model with two genetic algorithms}
\label{coevolution_schema}
\end{figure}

Provided we can formulate an adequate representation of i) the virus genome and ii) policy measures, and under the assumption that the mappings a)  between the virus genome and the virus phenotype and b) between the policy actions and the virus phenotype fitness, can be modelled in a satisfying way, we can represent their coevolution as a dual genetic algorithm with two populations: a population of viruses, and a population of policy measures. Both interact indirectly on a third population: the general human population. Viruses survive by infecting new humans in that population, and policy measures modify -to some extent- the behavior of the human population, as Figure \ref{coevolution_schema} illustrates.
    
Why GAs? Genetic algorithms are relevant tools to model this coevolution relationships for several reasons. First, evolutionary algorithms appear relevant to model natural selection contexts, as this is precisely their main focus (Holland 1992), though a significant fraction of the literature has used this method for optimisation. Second, among evolutionary algorithms, the inner genetic-centered approach of GAs give them an adequate baseline to encode more complex genomes and phenotypes. The computational architecture of GAs centered on a genetic representation, subject to evolution operators, appears to be the closest to the biological objects we are here interested in modelling. Third, genetic algorithms are particularly powerful in exploring new regions of large search spaces (Whitley, 1994), that may have non trivial structure (Wiransky, 2020). In our coevolution context, we are interested to see what new features may emerge from both viruses and policy responses. GAs, that can generate this novelty, thus constitute a relevant option. Fourth, coevolution has already been modelled using GAs for optimisation (Potter and De Jong, 1994, Vie, 2020b), giving solid foundations for further work in the area, and existing tools to understand the complex dynamics of the artificial SARS-CoV-2 coevolution. \\
    
How could this artificial coevolution be implemented? Starting from initial conditions constituted by i) a population distribution of SARS-CoV-2 variants with identified genome sequences and traits and ii) a distribution of the current policy measures, we can simulate the evolution of viruses and policy actions, in response one to another. \par
To define fitness in this world, one could assume that viruses simply aim at surviving, and do not have an objective function defining some metric to maximise; the performance of policy measures could be evaluated by minimising the number of deaths or infections. \par
The source of novelty in this coevolution system would essentially be mutations for viruses, and both mutations and recombination for policy measures. While viruses infect new hosts, and don't reproduce between themselves, it is reasonable to consider that national policy makers are exchanging, taking note of what happened in other countries, and changing their own actions in response to positive effects. \par
From this starting condition, and under these evolution criteria and mechanisms, a large number of runs of the system could be simulated. By observing the behavior of the artificial viruses and policies, and the outbreak dynamics in the artificial human population, some insights could emerge. We could discover some regularities, such as seeing whether and when viruses evolve towards greater transmsissibility, but also observe the changes in the genome, providing useful indications on where to experimentally look at during physical genome sequencing. \par
This artificial coevolution system may offer us a laboratory to "debug" our current policy measures, identify the weaknesses of our current strategies, and anticipate the evolution of the virus. If a significant portion of the simulations produced viruses that find a way to not be detected by PCR tests, or to evolve a resistance to our current vaccines, policy makers could be advised in advance of this possibility, and work ahead to prevent this issue from happening. \\
    
At times where policy makers faced significant uncertainty on the impact of their measures, a difficulty exacerbated by the rather long incubation time of SARS-CoV-2 (Lei et al., 2020), this artificial coevolution system can provide them with a complementary way to assess the impact of prospective policy measures, with an emphasis given on the evolution of the virus. In other words, such simulation possibilities may give the policy maker not only an estimate of the impact of the measures over infection rates and death rates, but also the possibility to consider the consequences of such measures over the future possible traits of the virus.  \par 

\section{An example of implementation}
\label{example}

In this section, we present an implementation example of a coevolution model with dual genetic algorithms. We highlight the building blocks of the model, the parameter configuration, and the key results.

\subsection{Model}

\paragraph{Genetic representation}

Individual viruses' genomes in the model are represented as a binary string whose length is the \textit{virus size}. Viruses are initialised with a genome composed exclusively of zeros: this assumes that at the start, viruses are an original form of the disease with no mutations. Each element of this genome represents activation (if equal to 1) or non-activation (if equal to 0) of specific mutated genes. Each mutated gene has an effect on the virus reproduction rate. These effects are drawn uniformly in the interval [-1,1]. This means that some mutations will be detrimental to the virus reproduction, others will have very small or null effects, and some will favor reproduction. We simplify as such the process and effects of mutations, collapsing all these dimensions onto the virus reproduction rate. The virus population contains a given number at the start, programmed by the parameter \textit{initial virus size}.

\begin{figure}
    \centering
    \includegraphics[width = \textwidth]{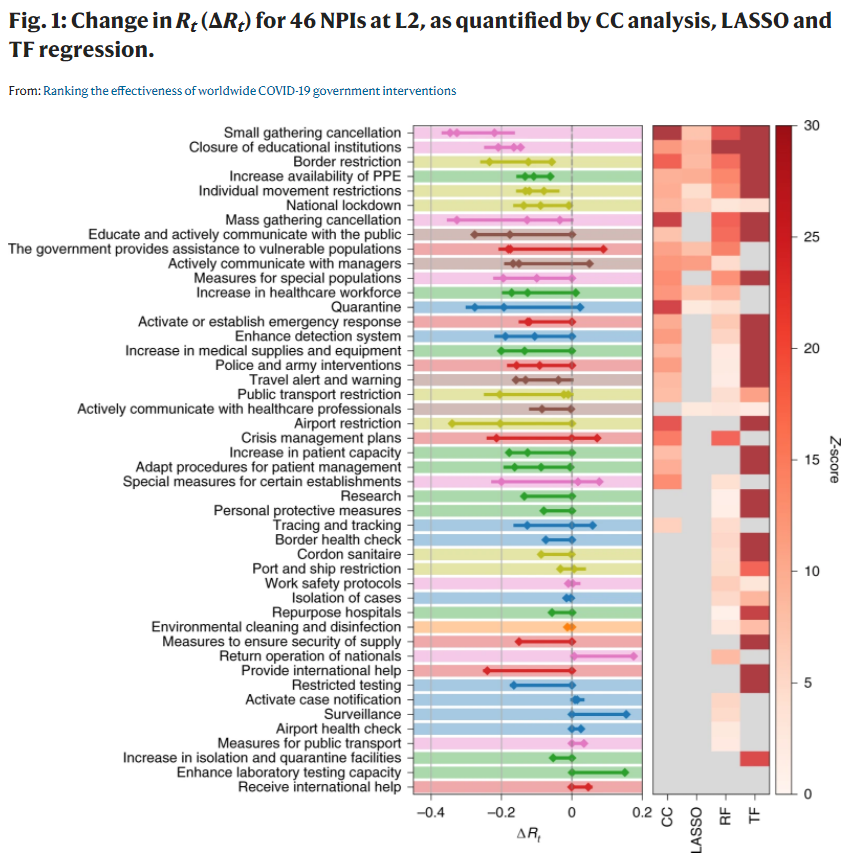}
    \caption{Effects of Covid 19 government interventions (From Haug et al., 2020). With permission from Nature Human Behavior - Reproduction License 4994130245697 (Jan. 22 2021)}
    \label{npi_effects}
\end{figure}

Individual policies are represented as a binary string as well, initialised with only zeros. This illustrates a starting point in which government policies start with no measure at all. Each element of the policy genome is a policy that can be activated (for a value of the corresponding genome location to 1). Again, we restrict our attention on the virus reproduction rate, and ignore all other dimensions. Each measure will have an effect over the virus reproduction rate, illustrating the efficiency of different measures to prevent the spread of the disease. The effects of these measures are calibrated from the values obtained by Haug et al. (2020) in their influential analysis of the impact of non pharmaceutical interventions. Our model captures the uncertainty on the effects of these policies by setting the effect to be drawn uniformly from the 95\% confidence intervals identified by Haug et al. (2020), illustrated in Figure \ref{npi_effects}. This draw is done once at the beginning of the run. The number of policies considered is parametrised with the \textit{policy population size} parameter, and will remain constant during the run. Policies can include up to 46 measures, corresponding to the measures studied by the above reference.

\paragraph{Infection process}

We adopt in this illustration a very simplified model of infection. Each individual virus in the population is characterised by a reproduction rate that incorporates two elements. First, a \textit{"base" reproduction rate}, corresponding to the reproduction rate of the original SARS-CoV-2. Second, this base rate is added to the sum of the effects of mutations activated by this particular individual virus' genome. In the infection step, each virus will infect as many hosts as its \textit{effective reproduction rate}. This effective reproduction rate is equal to the virus reproduction rate, minus the average reduction in reproduction rate in the policy population. \par

For each new infection, random mutations will happen with a given probability: the \textit{virus mutation rate}. Each element of the virus genome can mutate independently. Higher mutation rates will lead the virus to mutate more frequently during infections. The mutation operator will transform the given element of the genome to a 1 if it is characterised by the value 0, and inversely. As a result, and as the pandemic grows or diminishes, the size of the population of viruses handled by the genetic algorithm will vary, and some diversity may appear within this population.

\paragraph{Fitness}

In this model, we reduce the decision makers' problem to a minimisation of the reproduction rate of the virus, which essentially encompass objectives of reduction of deaths. Each individual policy is characterised by a total reduction in the reproduction rate, equal to the weighted sum of the effects of the activated specific measures. The fitness, or value of each individual policy, will evaluate the weighted effective reproduction rate of three viruses chosen at random in the virus population, in a tournament selection process. The policy reduction in the reproduction rate will be applied, and the net, effective reproduction rate recorded. Policies that obtain lower effective reproduction rates will be more likely to be selected in the creation of the next generation of policies.\par

Viruses do not mutate with an objective. Hence, we have not included a fitness function for the evolution of viruses. Mutations remain unguided by any objectives. The changes of the population of viruses will be driven by the differential reproduction rates of various strains, as described below.

\paragraph{Policy learning}

After the fitness of the policies has been determined, policies will be selected to form the basis of next generation policies using "roulette wheel" cumulative fitness selection. Each policy's selection probability will be equal to the ratio of its adjusted fitness (equal to $\frac{1}{1+r}$ where $r$ is the effective reproduction rate of the policy) to the sum of adjusted fitness scores. This crossover step models a process of communication between successful policies: decision makers observe their peers in other countries, observe the measures they implement and the associated results. Measures that appear efficient abroad tend to be implemented nationally by the means of this imitation step. This crossover step occurs with probability equal to the \textit{policy crossover rate}. After selecting two policies, a random uniform crossover point will be determined, and the two policies' genomes will be interchanged after this crossover point. The result of this procedure will be two children policies for the next generation. Otherwise, when the crossover operator is not activated with probability 1 - \textit{policy crossover rate}, the children strategies will be exact copies of their parents.\par

Learning to improve policies will also include a mutation step, modelling small perturbations or explorations. This illustrates for instance a country implementing or removing quarantine restrictions for various reasons. With a \textit{policy mutation probability}, any element can mutate from value 0 to value 1. We outline here one important limitation: we do not allow policies in our model to revert back after some measures have been implemented: we essentially forbid detrimental mutations. Extending our space of possible measures to measures that do not work could be an interesting direction as well. We also do not consider other factors such as economic output or political situation that could act as a pressure towards relaxation of measures. Again, these constraints would be an interesting addition for this model, but we have chosen to present a simple illustration of coevolution.

\paragraph{Evolution run and parameters}

The simulation runs for \textit{Tmax} periods. We run our simulations for a base reproduction rate of 2.63 (Mahase, 2020). Note however that simply changing the value of the base reproduction rate, or including uncertainty on its determination, is easily achievable in the source code (see below for availability). Higher base rates will likely make the infection spike faster and higher, while lower base rates may lead to the virus extinction in some cases, or reductions of the outbreak peaks. In the model, we consider the time periods to be indexed as weeks, assuming that each virus is transmitted every seven days.

\begin{table}[H]
    \centering
    \begin{tabular}{|c|c|}
    \hline
        Parameter & Value \\
        \hline
        Virus initial population size & 10 \\
        Virus size & 10 \\
        Policy population size & 100 \\
        Base reproduction rate & 2.63 \\
        Tmax & 20 \\
        Policy crossover rate & 0.5 \\
        Policy mutation rate & 0.05 \\ 
        Virus mutation rate & 0.0001 \\
        \hline
    
    \end{tabular}
    \vspace{0.3cm}
    \caption{Parameter configuration for the dual genetic algorithm}
    \label{parameters}
\end{table}

A situation of coevolution defines a run in which both the viruses and the policy can evolve: that is, their mutation rates and the policy crossover rate are strictly positive. When the virus mutation rate is null, but the policy mutation rate and policy crossover rates are positive, we model a situation in which only the policy is evolving, against a static virus. When the virus mutation rate is positive, and the policy mutation rate and crossover rates are null, we are illustrating a situation in which the virus evolves, and policies remain indifferent and void. All other parameters remain unchanged.

Before turning to the simulation results, we make a note on the impact of the parameters over the results, and the outbreak dynamics that are generated. A major challenge in this example implementation was to avoid too large epidemics: as each virus is simulated individually, handling hundred of millions of viruses can incur a significant computational cost. The development of the simulation allowed us to be able to simulate in reasonable time (seconds) up to ten billion individual viruses. 
Higher virus mutation rates, or higher initial virus sizes, or less effective policies, can lead to exponential growth of the virus population size.  Alternatively, if policies are very efficient (high mutation rates and crossover rates), and if the virus does not mutate frequently enough, the model may manage to make the virus go extinct. We must acknowledge that simulation results can be sensitive to small variations of the parameters. The configuration showed in Table \ref{parameters} allows to keep computation doable for the 20 time periods considered. Outside extreme situations (complete virus takeover or virus extinction), the main insights presented below hold.

\subsection{Results}

\begin{figure}[H]
        \centering
        \begin{subfigure}[b]{0.495\textwidth}
            \centering
            \includegraphics[width=\textwidth]{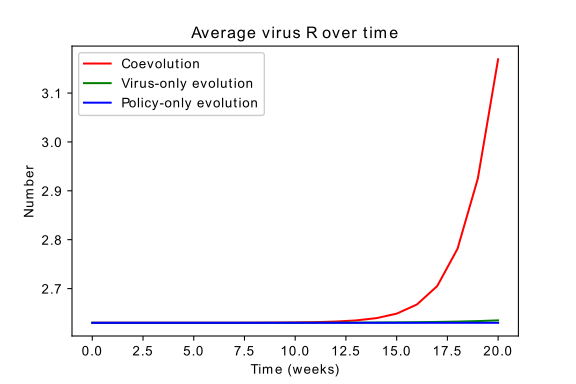}
            \caption[]
            {{\small Average reproduction rate of the population of viruses over time}}   
            \label{virusR}
        \end{subfigure}
        \hfill
        \begin{subfigure}[b]{0.495\textwidth}  
            \centering 
            \includegraphics[width=\textwidth]{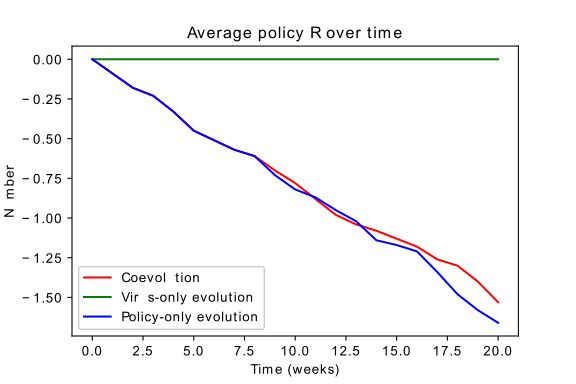}
            \caption[]
            {{\small Average impact in reproduction rate of policies over time}} 
            \label{policyR}
        \end{subfigure}
        \vskip\baselineskip
        \begin{subfigure}[b]{0.495\textwidth}   
            \centering 
            \includegraphics[width=\textwidth]{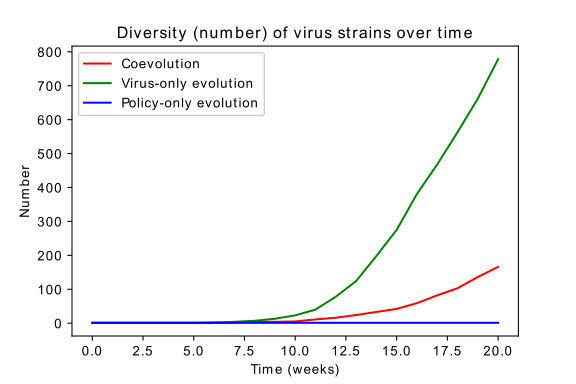}
            \caption[]
            {{\small Number of different virus strains over time}} 
            \label{diversity}
        \end{subfigure}
        \hfill
        \begin{subfigure}[b]{0.495\textwidth}   
            \centering 
            \includegraphics[width=\textwidth]{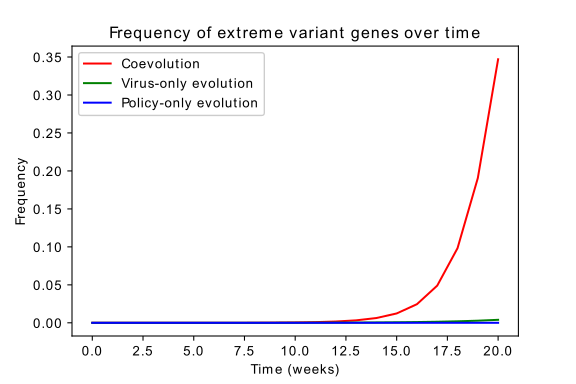}
            \caption[]
            {{\small Frequency of extreme variant genes over time}} 
            \label{virus_genome}
        \end{subfigure}
        \caption[Key results from the coevolution dual genetic algorithm]
        {\small Key results from the coevolution dual genetic algorithm} 
        \label{results}
    \end{figure}

We now run the evolution of viruses and policies in these three situations above, to identify specific features of the coevolution regime. The Figure panel \ref{results} presents the main results. Their observation allows us to formulate a few "stylized facts" of the coevolution of viruses and policies.

\begin{enumerate}
\setlength\itemsep{1em}
    \item \textbf{Under coevolution, virus adaptation towards more infectious variants is considerably faster than when the virus evolves against a static policy. }
    In Figure \ref{virusR}, we can observe that the average reproduction create in the virus population rises to 3.1 after 20 time periods under coevolution (red curve). When the virus does not evolve (blue), the average reproduction rate naturally stays at the initial value of 2.63. Interestingly, when the virus can evolve, but when the policy does not (green curve), the average reproduction rate tends to increase slightly, but much less than under the coevolution regime. Having the virus face a more severe struggle for its survival makes its evolution more efficient.
    
    \item \textbf{More contagious strains become dominant in the virus population under coevolution}.
    Figure \ref{virus_genome} shows the frequency of viruses in the virus population containing the mutation gene granting the highest increase in reproduction rate. This fraction rises to 0.35 in the coevolution case, while this share is considerably lower under virus-only evolution. This point supports the idea that coevolution makes virus' adaptation much more efficient. Indeed, the number of different variants in the population exposed by Figure \ref{diversity} shows interesting insights. In the virus-only evolution, up to 800 variants appear during the 20 time periods. This is due to the outbreak dynamic: in the virus-only evolution, policies do not do anything and do not change, hence the virus is free to spread everywhere. As its population size grows, more mutations happen, and more variants emerge. Under coevolution, only up to 200 variants emerge, but the frequency of the strongest mutations shows that virus evolution is made much more efficient by the challenge proposed by learning policies. 
    
    \begin{figure}[H]
        \centering
        \includegraphics[scale = 0.65]{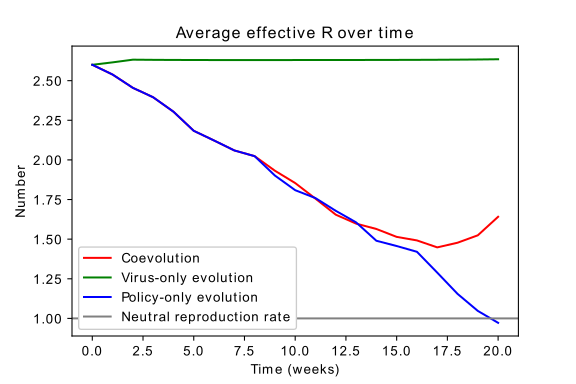}
        \caption{Average effective reproduction rate over time}
        \label{effectiveR}
    \end{figure}
    
    \item \textbf{The coevolution regime can generate multiple outbreaks waves as the more infectious variants becoming more dominant in the virus population}.
    While currently in European countries, a so-called third wave seem to have occurred coincidentally to the VOC 202012/01 (the "UK variant") becoming dominant, this pattern occurred as well during our evolution run. Figure \ref{policyR} shows that policies evolve to be more efficient over time, leading the average effective reproduction rate of the virus to go below 1, in a path to extinction. Under the coevolution regime, the more efficient adaptation of the virus allows instead the effective reproduction rate to increase again. Several multiple waves seem empirically to stem from relaxing measures, a behavior that our model does not include. However, the same pattern and insight would hold. In this simulation of coevolution, multiples waves of infection can occur because of increasing viruses' reproduction rates, or relation of policy measures.  
    
    \item \textbf{Seeing more infectious virus variants becoming dominants may signify that our policy measures are effective}. 
    These sets of figures show that when policies are not evolving and not effective, more infectious variants take a much longer time to become dominant in the population. Only when policies evolve and actively undermine the virus reproduction, weaker forms progressively disappear, to be replaced by stronger virus variants. Several countries today see numerous variants quickly increase in the share of new infections. While this dynamic constitutes a key challenge and difficulty, it can be seen as the sign that the current measures are putting stress on the virus: they are efficient in pushing weaker forms to reduction and eventually extinction. Only by continuously adapting, and adapting faster than the virus strains, can policies and human behaviors push all variants to final extinction. Our future work with this model will strive to include vaccines as a policy measures, allow viruses to obtain a vaccine-resistant trait by mutations, and observe how the evolution of policies shapes the emergence of vaccine-resistant strains of SARS-CoV-2.  
\end{enumerate}

\section{Implications and perspectives}
\label{perspectives}

This perspective for the artificial coevolution first faces the challenges inherent to the use of GAs, that were recently reviewed by Vie (2020a). Their computational cost increases significantly with the size of the populations they consider. If we wanted to simulate very large population of viruses, knowing that the evolution of SARS-CoV-2 is a hugely parallel process occurring over millions of hosts simultaneously, the computational cost of the simulation would be significant. In addition, small differences in parameter configuration of GAs, including population size, mutation rates, selection intensity, is difficult in GAs, as different sets of parameters may yield different results, and impact the algorithm performance, or convergence properties (Grefenstette, 1986). Last but not least, the genetic representation needs careful design to cover the diversity of possible solutions in a realistic manner, without creating unintended loopholes that could be exploited by the algorithm (Juzonis et al., 2012) and bias the results. Several recent works shed new light on these challenges, and provide new means to mitigate their effects. The computational cost of GAs fades before their great scaling with parallelism (Mitchell, 1988), and the computing power of GPUs (Cheng and Gen, 2019) or Cloud computing hardware. New methods have been introduced in parameter configuration (Hansen, 2016; Huang et al., 2019; Case and Lehre, 2020). A large diversity of genetic representations exist in GAs, and some further inspiration from key biological concepts can open the way to representations allowing these algorithms to evolve more complex artificial organisms (Miikkulainen and Forrest, 2021).\\ 
    
Specifically in the perspective of the artificial coevolution laboratories discussed here, a key challenge remains in establishing a proper algorithmic representation of the SARS-CoV-2 genome, and the mapping between this genome and the virus traits. By proper, we mean that this representation might not need to be comprehensive or perfectly exact, but should not oversimplify the object being studied, or neglect important determinants of traits. The work perspective described here faces important limitations, and as these algorithms could be used for essential matters of public health, the biases they may contain require careful consideration. These programs cannot simulate at perfection natural selection or comprehensive genetics, simply because we do not fully understand them yet.\\ 
    
Attempting to model the coevolution of viruses with more realistic simulations than the example provided here is certainly a challenging endeavor. It however entails significant benefits and opportunities. The recent mutations of SARS-CoV-2 have raised public awareness about this critical issue for public health, and make attempts to address this issue with a matter of public interest, with immense benefits when we consider the cost faced by the general public due to variants-caused restrictions. This challenge constitutes as well an opportunity for evolutionary algorithms to grow. If we can make these computer programs that simulate natural selection capable of representing and simulating the evolution of viruses, which are organisms considerably more complex that what EAs are currently handling, these improved EAs in the future could lead to breakthroughs in bioinformatics, optimisation, artificial life and AI. \par
How could such algorithms evolve organisms with that level of complexity? Modifications of GAs that move from the simple bit string representation to more complex genomes, can start this transformation. Key phenomena in genetics and biology such as \textit{pleiotropy} -where one gene impacts several traits-,\textit{polygeny} -one trait is impacted by several genes-, the evolution of \textit{evolvability}, realistic mutations, are yet to be included in these algorithms, and their addition carries significant benefits and new opportunities. These "structural" genetic algorithms that place such emphasis on the genome structure, may make us able to evolve much more complex, adaptive artificial entities to study viruses evolution as illustrated here, but also to create advanced forms of artificial life, or foster progress in generative artificial intelligence. The challenge of modelling SARS-CoV-2 coevolution with genetic methods can inspire such decisive innovations.
    
\section{Data and code availability}
\label{data}

The main simulation code of the GA proof of concept is freely available at \url{https://github.com/aymericvie/Covid19_coevolution}. Model parameters such as the efficiency of different non pharmaceutical interventions, or the basic reproduction rate of SARS-CoV-2, as well as mutation rates, or learning rates for policies, can be easily changed in the code. The code is designed to work on Google Colab, and the script is self sufficient to run.

\section{Conclusion}
\label{conclusion}
In this article, we propose coevolution with genetic algorithms (GAs) as a credible approach to model this relationship, highlighting its implications, potential and challenges. We provide a proof of concept-implementation of this coevolution dual-GA.  Because of their qualities of exploration of large spaces of possible solutions, capacity to generate novelty, and natural genetic focus, GAs are relevant for this issue. We present a dual GA model in which both viruses aiming for survival and policy measures aiming at minimising infection rates in the population, competitively evolve. Under coevolution, virus adaptation towards more infectious variants appear considerably faster than when the virus evolves against a static policy. More contagious strains become dominant in the virus population under coevolution. The coevolution regime can generate multiple outbreaks waves as the more infectious variants becoming more dominant in the virus population. Seeing  more  infectious  virus  variants  becoming  dominants  may signify  that  our  policy  measures  are  effective. This artificial coevolution system may offer us a laboratory to "debug" our current policy measures, identify the weaknesses of our current strategies, and anticipate the evolution of the virus to plan ahead relevant policies. It also constitutes a decisive opportunity to develop new genetic algorithms capable of simulating much more complex objects. We highlight some structural innovations for GAs for that virus evolution context that may carry promising developments in evolutionary computation, artificial life and AI.

\end{document}